\documentclass[conference]{IEEEtran}
\IEEEoverridecommandlockouts
\usepackage{cite}
\usepackage{amsmath,amssymb,amsfonts}
\usepackage{graphicx}
\usepackage{textcomp}
\usepackage{xcolor}
\usepackage{array}
\usepackage{subfigure}
\usepackage{verbatim}
\usepackage{enumitem}
\usepackage{booktabs}
\usepackage{multirow}
\usepackage{algorithm}
\usepackage{algpseudocode}
\usepackage{graphicx}
\usepackage{balance}

\newcommand{\DeepMorph}{{\small{\textsf{DeepMorph}}}} 
\newcommand{\TensorFlow}{{\small{\textsf{TensorFlow}}}} 
\newcommand{\MNIST}{{\small{\textsf{MNIST}}}} 
\newcommand{\cifar}{{\small{\textsf{Cifar-10}}}}

\newcommand{\SD}{{\small{\textsf{SD}}}}  
\newcommand{\ITD}{{\small{\textsf{ITD}}}} 
\newcommand{\UTD}{{\small{\textsf{UTD}}}} 
 
\newcommand{\softmax}{{\small{\textsf{softmax}}}}

\def\BibTeX{{\rm B\kern-.05em{\sc i\kern-.025em b}\kern-.08em
    T\kern-.1667em\lower.7ex\hbox{E}\kern-.125emX}}
\begin{document}
\balance

\title{Detecting Deep Neural Network Defects with Data Flow Analysis}
\author{
    \IEEEauthorblockN{Jiazhen Gu\IEEEauthorrefmark{1}\IEEEauthorrefmark{2}, Huanlin Xu\IEEEauthorrefmark{1}\IEEEauthorrefmark{2}, Yangfan Zhou\IEEEauthorrefmark{1}\IEEEauthorrefmark{2}, Xin Wang\IEEEauthorrefmark{1}\IEEEauthorrefmark{2}, Hui Xu\IEEEauthorrefmark{3}, Michael Lyu\IEEEauthorrefmark{3}}
    \IEEEauthorblockA{\IEEEauthorrefmark{1}School of Computer Science, Fudan University, Shanghai, China
    \\\IEEEauthorrefmark{2}Shanghai Key Laboratory of Intelligent Information Processing, Shanghai, China
    \\\IEEEauthorrefmark{3}The Chinese University of Hong Kong
    }
}

\maketitle
\begin{abstract}
Deep neural networks (DNNs) are shown to be promising solutions in many challenging artificial intelligence tasks. However, it is very hard to figure out whether the low precision of a DNN model is an inevitable result, or caused by defects. This paper aims at addressing this challenging problem. We find that the internal data flow footprints of a DNN model can provide insights to locate the root cause effectively. We develop DeepMorph (DNN Tomography) to analyze the root cause, which can guide a DNN developer to improve the model.
\end{abstract}

\section{introduction}
\label{sec:intro}
\begin{table*}[htb]
    \centering
    \caption{Results on DL models with injected defects}
    \resizebox{\textwidth}{!}{
    \begin{tabular}{|c|c|c|c|c|c|c|c|c|c|c|c|c|}
        \toprule
        \multirow{3}{*}{\textbf{Defect}} & \multicolumn{6}{c|}{\textbf{MNIST}} & \multicolumn{6}{c|}{\textbf{CIFAR-10}} \\
        ~  & \multicolumn{3}{c|}{\textbf{LeNet}} & \multicolumn{3}{c|}{\textbf{AlexNet}} & \multicolumn{3}{c|}{\textbf{ResNet}} & \multicolumn{3}{c|}{\textbf{DenseNet}} \\
        ~  & ITD & UTD & SD & ITD & UTD & SD & ITD & UTD & SD & ITD & UTD & SD \\
        \midrule
        \textbf{ITD} & {\color{red}\textbf{0.763}} & 0.011 & 0.226 & {\color{red}\textbf{0.822}} & 0.023 & 0.155 & {\color{red}\textbf{0.694}} & 0.234 & 0.072 & {\color{red}\textbf{0.770}} & 0.191 & 0.039 \\
        \midrule
        \textbf{UTD} & 0.152 & {\color{red}\textbf{0.745}} & 0.103 & 0.145 & {\color{red}\textbf{0.787}} & 0.068 & 0.138 & {\color{red}\textbf{0.577}} & 0.285 & 0.185 & {\color{red}\textbf{0.643}} & 0.172 \\
        \midrule
        \textbf{SD} & 0.280 & 0.091 & {\color{red}\textbf{0.629}} & 0.238 & 0.174 & {\color{red}\textbf{0.588}} & 0.433 & 0.086 & {\color{red}\textbf{0.481}} & 0.452 & 0.013 & {\color{red}\textbf{0.535}} \\
        \bottomrule
    \end{tabular}}
    \label{exp:inj_def}
\end{table*}
Machine learning with deep neural networks ({\em i.e.}, deep learning, or DL) has been proven effective in many specific tasks~\cite{cvpr/HeZRS16, interspeech/SundermeyerSN12, corr/BahdanauCB14}, and even those in traditional domains~\cite{misc/ai_bio}.
Developers tend to use modern DL software frameworks, {\em e.g.}, TensorFlow~\cite{osdi/AbadiBCCDDDGIIK16}, to develop DL applications even without much professional DL knowledge~\cite{misc/tf_image}.

The implementation of DL application rarely introduces traditional software bugs, while DL applications still frequently exhibit undesirable behaviors, leading to unexpected mistakes~\cite{misc/AiIgnore}. 
When a DL model behaves unexpectedly, it is hard for developers to localize defects in the model.

This paper presents \DeepMorph\ (Deep neural networks Tomography), a tool to detect defects in DL models. We model the execution process of DL models through extracting internal data flow footprints, {\em i.e.}, the intermediate outputs of every layers. We find that such footprints can provide insights to locate the root cause effectively, which can instantly direct a developer to improving the DL model. 

Our contributions of this work are highlighted as follows.
\begin{enumerate}
    \item We systematically analyze the execution process of DL models in a software engineering perspective. We interpret how the intermediate outputs of the hidden layers describe the execution process of a DL model.
    \item We present a new method to extract the internal data flow footprints, which can be used to reason bad performance and locating the corresponding defects.
    \item We implement \DeepMorph\ based on \TensorFlow, a widely-adopted DL software framework. We show the effectiveness of \DeepMorph\ in locating model defects with different DL models trained on four popular datasets.  
\end{enumerate}

The rest of this paper is organized as follows. We introduce related work in Section~\ref{sec:related}. 
We elaborate our \DeepMorph\ design in Section~\ref{sec:method}. Section~\ref{sec:eval} provides the details of our experimental study. We provide further discussions and conclude the work in Section~\ref{sec:con}.

\section{related work}
\label{sec:related}

In recent years, machine learning with deep neural networks has surged into popularity in many application areas~\cite{cvpr/HeZRS16, interspeech/SundermeyerSN12, corr/BahdanauCB14}, including solving traditional software engineering problems~\cite{dsn/LeeL17, dsn/KimHOL18, icse/ChenSMXL18, icse/GuZ018, dsn/NieXGPEST18}. 

Meanwhile, with the wide adoption of machine learning based applications, the reliability and security of machine learning-based software also attract much research attention~\cite{dsnw/AntunesBFLMS18, dsnw/GhoshJTL018, dsnw/LuCCY18, dsnw/WenHYZ18}. 
This line of work mainly focuses on understanding and addressing inherent defects in DL models. Our work, in contrast, intends to locate model defects introduced by improper network design or faulty training data, and direct developers to further improve their DL models.

Moreover, various methods have been proposed for testing and debugging DL models~\cite{sosp/PeiCYJ17, icse/TianPJR18, kbse/MaJZSXLCSLLZW18, kbse/SunWRHKK18, kbse/ZhangZZ0K18, issta/DwarakanathASRB18, issre/MaZSXLJXLLZW18, ma2018mode, corr/abs-1803-04792, corr/abs-1807-10875, ijcai/RuanHK18, corr/abs-1710-00486}.
In this paper, we present \DeepMorph~to summarize typical model defects through analyzing the data flow footprints inside DNN models.
In contrast to existing approaches, \DeepMorph~pays attention to locate the root cause of bad model performance, instead of aiming solely on performance improvement.

\section{Methodology}
\label{sec:method}

\DeepMorph~is a tool designed to facilitate the developer to analyze whether there is a potential defect that causes the bad performance of the model.
It is promising to locate the root cause of bad model performance via analyzing the internal data flow footprints. 
This section illustrates how we realize an automatic approach to this end. 

In this paper, we focus on three representative types of model defects as follows.
\begin{itemize}
\item {\em Structure Defect} (\SD): The improper network structure leads the model to learn inappropriate features from the training data.
\item {\em Insufficient Training Data} (\ITD): The distribution of the training data is different from that of the data in production environment.  
\item {\em Unreliable Training Data} (\UTD): The training set contain falsely labeled cases.
\end{itemize}

\begin{figure}[t!]
\centering
\includegraphics[width=\linewidth]{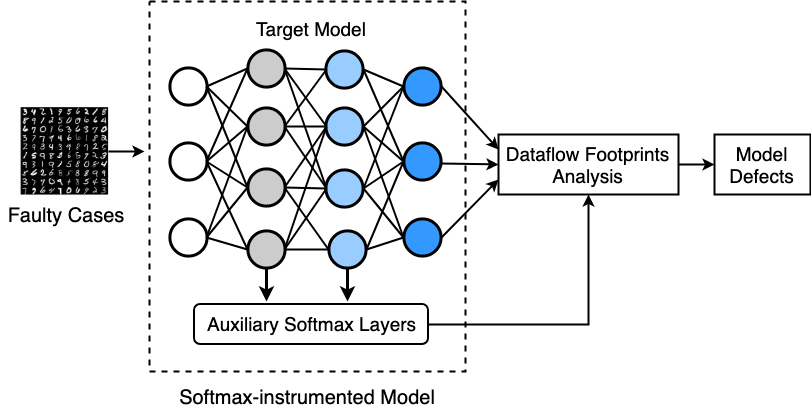}
\caption{Overview of DeepMorph}
\label{fig:overview}
\end{figure}

Figure~\ref{fig:overview} illustrates the design of \DeepMorph. 
When the performance of a deep neural network is lower than expected, 
\DeepMorph~first builds the \softmax-instrumented model via adding auxiliary \softmax~layers to the target model.
The \softmax-instrumented model is used to learn the execution pattern of the training cases for each target class. 

Then \DeepMorph\ feeds the faulty cases to the \softmax-instrumented model, which extracts data flow footprint specifics from the intermediate outputs of hidden layers in the target model.

The footprint specifics are capable of representing the classification process, and allow \DeepMorph\ to compare the footprints against the execution pattern of each target class. 
By examining the process, layer by layer, of how inputs are misclassified, \DeepMorph\ can then reason the defect that causes the faulty cases. 

\section{experiments}
\label{sec:eval}

We implement \DeepMorph\ over \TensorFlow, a widely-adopted DL software framework. 
Our experiments are designed with a focus on answering how effective \DeepMorph\ is in locating model defects in controlled environments.
Based on the defect reported by \DeepMorph, we modify the models accordingly and evaluate whether \DeepMorph\ is helpful to improving model performance.

We train the corresponding DL model with a set of training data.
The DL model is applied to the test data used to emulate those encountered in production environments.
Given the target DL model, training set, and faulty cases found in the test data, \DeepMorph\ first builds the \softmax-instrumented model and trains the auxiliary \softmax~layers with the training data.
The \softmax-instrumented model processes the faulty cases and produces the footprint specifics of these cases.
Then for all faulty cases, it produces the ratio of each type of defects.
The defect with the highest ratio value is considered to be the dominant defect of the target DL model.

We employ two standard datasets for image classification: \MNIST\ and \cifar~\cite{misc/cifar}, both of which have 10 target classes labeled as 0-9.
We consider 4 typical DL classifier implementations from Github~\cite{misc/cifarzoo}.
For \MNIST, we utilize LeNet~\cite{nips/CunBDHHHJ89} and AlexNet~\cite{krizhevsky2012imagenet}, which have 5 and 8 layers respectively.
For \cifar, we use ResNet-34~\cite{cvpr/HeZRS16} and DenseNet-40~\cite{cvpr/HuangLMW17}.

To study how DeepMorph performs in locating each defect, we manually inject the defects to these DL-models and conduct our experimental study, which is elaborated as follows.

\begin{itemize}
    \item \ITD\ happens when the data distribution have obvious difference between training data and test data. We randomly remove a part of data of some specific classes.
    \item \UTD\ refers to the unreliable training data and happens when human made mistakes. We tag a part of the training data of one class to the other.
    \item We inject \SD\ through manually removing three kinds of layers, namely Convolution layer from the original network structures, which aims at degrading the models via a weaker network structure. 
\end{itemize}

The results reported by \DeepMorph\ on models with different injected defects are shown in Table~\ref{exp:inj_def}.
We can see that, for all cases, \DeepMorph\ is able to locate the injected defect effectively. 
According to the ratio values reported by \DeepMorph, the injected defects are always the largest.
This indicates that \DeepMorph\ can successfully identified the injected defects.

\section{Conclusion}
\label{sec:con}

This paper aims at addressing the model defects of DL applications.
We argue that the model defects, {\em i.e.}, those caused by improper network structure and improper network parameters (caused originally by improper training data), can be located with a white-box approach. 

We formulate a DL model as a functional composition of hidden layers, and analyze its execution with data flow footprints.
We attempt to interpret the model execution as how the distinct features of an input case towards the DL task can be extracted layer by layer gradually. 

We demonstrate the effectiveness of our proposal by implementing a tool, namely, \DeepMorph. The results show it is very promising for \DeepMorph\ in locating model defects. Moreover, it can greatly facilitate DL model developers in guiding them towards improving the model.
\newpage

\bibliographystyle{IEEEtran}
\bibliography{dl_testing}    

\end{document}